\documentclass[10pt,twocolumn,letterpaper]{article}

\usepackage{cvpr}      
\usepackage{multirow}
\usepackage{graphicx}

\definecolor{cvprblue}{rgb}{0.21,0.49,0.74}
\usepackage[pagebackref,breaklinks,colorlinks,allcolors=cvprblue]{hyperref}

\def\paperID{2377} 
\def\confName{CVPR}
\def\confYear{2025}

\title{Learning Dynamic Collaborative Network for Semi-supervised 3D Vessel Segmentation}

\author{
Jiao Xu$^{1}$\thanks{Equal contribution. The majority of this work was completed while Xin Chen was at Dalian University of Technology.} , Xin Chen$^{2}$\footnotemark[1] , Lihe Zhang$^{1}$\thanks{Corresponding author} \\
$^{1}$Dalian University of Technology  $^{2}$City University of Hong Kong \\
{\tt\small xjmmcome@mail.dlut.edu.cn, xche32@cityu.edu.hk, zhanglihe@dlut.edu.cn}
}

\begin{document}
\maketitle
\pagestyle{empty}
\thispagestyle{empty}
\begin{abstract}
In this paper, we present a new \textbf{d}ynam\textbf{i}c \textbf{co}llaborative network for semi-supervised 3D vessel segmentation, termed DiCo. Conventional mean teacher (MT) methods typically employ a static approach, where the roles of the teacher and student models are fixed. However, due to the complexity of 3D vessel data, the teacher model may not always outperform the student model, leading to cognitive biases that can limit performance. To address this issue, we propose a dynamic collaborative network that allows the two models to dynamically switch their teacher-student roles. 
Additionally, we introduce a multi-view integration module to capture various perspectives of the inputs, mirroring the way doctors conduct medical analysis. 
We also incorporate adversarial supervision to constrain the shape of the segmented vessels in unlabeled data. In this process, the 3D volume is projected into 2D views to mitigate the impact of label inconsistencies.
Experiments demonstrate that our DiCo method sets new state-of-the-art performance on three 3D vessel segmentation benchmarks. 
The code repository address is \href{https://github.com/xujiaommcome/DiCo}{https://github.com/xujiaommcome/DiCo}.
\end{abstract}    
\section{Introduction}
\label{sec:intro}
In clinical applications, high-quality vascular imaging is crucial for radiologists to accurately detect and diagnose lesions, which are often subtle and challenging to identify. Consequently, improving the accuracy of 3D vessel perception is a pressing need. However, segmenting 3D vascular structures presents two challenges: i) \textbf{Scarcity of labeled data}: Labeling vessels demands extensive expertise and is labor-intensive, resulting in limited availability of high-quality annotated data. ii) \textbf{Complex appearance}: Vessels exhibit significant variability in their topological structures and diameters. Their continuous, long, and thin nature complicates the task of maintaining segmentation continuity, in contrast to other organs with more stable shapes.

\begin{figure}[!t]
    \begin{center}
        \begin{subfigure}{1\linewidth}
            \centering
            \includegraphics[width=\linewidth]{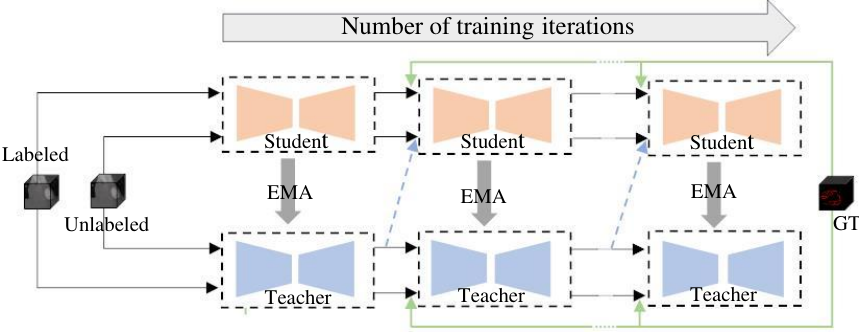}
            \subcaption{Static supervision}
            \label{fig:sub1}
        \end{subfigure}
        \begin{subfigure}{1\linewidth}
            \centering
            \includegraphics[width=\linewidth]{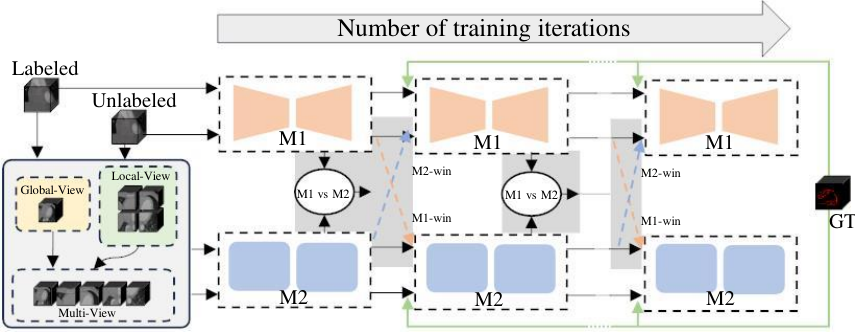}
            \subcaption{Our dynamic supervision}
            \label{fig:sub2}
        \end{subfigure}
    \end{center}
    \caption{Semi-supervised segmentation frameworks. The dotted arrows indicate the supervision information flow of unlabeled data.}
    
    \label{fig:pipeline}
\end{figure}

Many efforts have been made to address these challenges. To tackle the \textbf{scarcity of labeled data}, semi-supervised learning (SSL) has gained increasing attention for its ability to leverage large amounts of unlabeled data. Among SSL methods, mean teacher (MT) approaches~\cite{MT,UA_MT,Magicnet,ASE-Net} based on consistency regularization are representative, as shown in Fig.~\ref{fig:pipeline}(a). Although these methods have achieved significant success in segmenting common organs and tumors, they do not generalize well to vessel segmentation. This is primarily due to cognitive biases introduced by static supervision, compounded by the complexity of vessel scenes.
More specifically:
i) The pseudo labels provided by the static teacher model are not always superior to those of the student model, potentially misleading the student model's learning process.
ii) If the student network learns incorrect information during early training stage, this misinformation can be propagated to the teacher model through exponential moving average (EMA).
Thus, as iterations progress, errors are propagated and amplified between the MT networks, ultimately causing segmentation results to diverge from the ground truth.

To address these issues, we propose a dynamic collaborative network (DiCo) for the semi-supervised 3D vessel segmentation in this paper, as illustrated in Fig.~\ref{fig:pipeline}(b). By dynamically switching the teacher-student roles of two models according to their current performance, DiCo alleviates the error propagation and amplification caused by static supervision. The underlying intuition is that the better-performing model at any given time should guide another model, rather than maintaining a fixed teacher-student relationship. 
At each training step, we compute the supervised loss based on the labeled data, designating the model with the lower loss as the teacher, while the other becomes the student. The rationale behind this approach is that, in each iteration, the labeled and unlabeled data come from the same source. Therefore, the sub-network that performs well on the labeled data generally performs better on the unlabeled data as well. 

Representing the \textbf{complex appearance} of 3D vessels requires robust feature modeling capabilities.
For this purpose, some efforts~\cite{Deu-net,DeU-Net2.0,DCU-net,PointScatter,tree-structured-seg} have developed specialized layers to accommodate the tubular structures of vessels, enhancing the focus on key features.
These methods compute features globally and lack attention to essential local information.
In contrast, we design a straightforward multi-view integration module that captures both local vascular details and global image context. Specifically, we reorganize the original input volume to create an enhanced volume that considers both local and global views for model input.
To achieve well-shaped vessel segmentation, we employ an adversarial training to align the distribution of the predicted unlabeled mask with that of the real labeled mask. Since these masks do not correspond to the same image, our goal is to ensure similar shape styles rather than pixel-level alignment. To this end, we project the 3D masks into 2D space to avoid pixel-wise correspondence supervision.
Concretely, the labeled 3D volume and its ground-truth mask, as well as the unlabeled image and its predicted mask, are projected into 2D along the $z$ axis through maximum-intensity projection (MIP).
Then, the projected 2D images and masks are integrated into a discriminator for adversarial supervision, ensuring that the distribution of the unlabeled predicted masks more closely matches that of the labeled ground-truth masks, thereby preserving the continuity of the vessel structure.

Extensive experiments demonstrate the effectiveness of our DiCo method.
For instance, it achieves an 86.05\% DSC score on the recent vessel benchmark CAS2023~\cite{cas2023}, surpassing the previous best method, MagicNet~\cite{Magicnet}, by 2.28\%, and achieves results comparable to fully supervised DSCNet~\cite{DSCNet} using only 5\% of the labeled data.On the large-scale ImageCAS~\cite{ImageCAS} dataset, our method, DiCo, achieves state-of-the-art performance across three key metrics.
In summary, the contributions of this work are as follows:
\begin{itemize}
\item 
We introduce a dynamic collaborative network for semi-supervised 3D vessel segmentation. This approach enables the two models in the conventional MT framework to dynamically switch the teacher-student roles, providing a new view to alleviate the cognitive bias problem.
\item 
We propose a straightforward multi-view integration module that captures both local vascular details and global image context, leading to robust appearance modeling that supports precise vessel segmentation.
\item 
We propose an MIP-based adversarial supervision that aligns the shape styles of predicted vessel masks with that of real mask labels, thereby enhancing the quality of predicted vessel masks.
\end{itemize}

\section{Related Work}
\label{sec:related-work}

\begin{figure*}[t]
\centering
\includegraphics[width=0.90\textwidth]{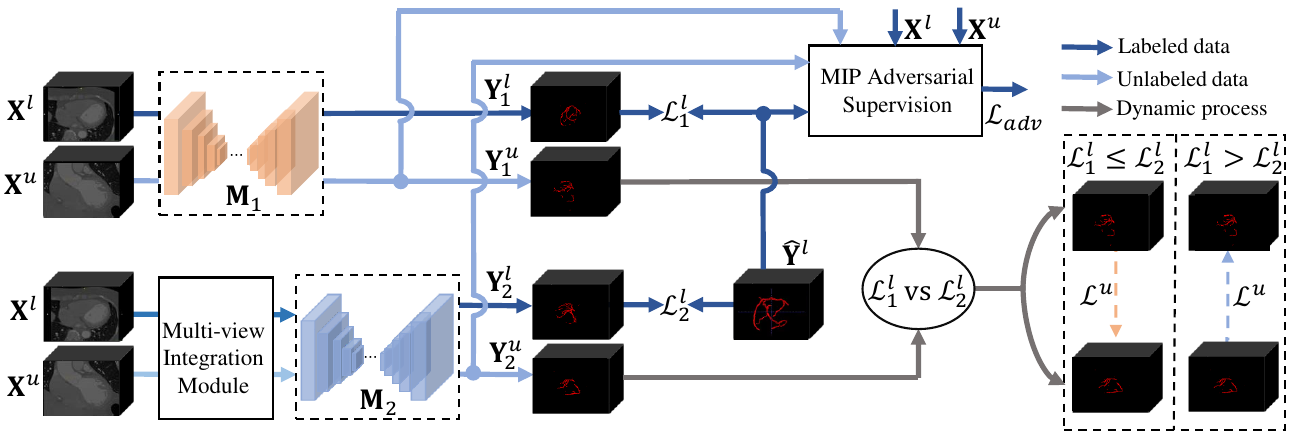} 
\vspace{-2mm}
\caption{Architecture of the proposed DiCo method. It consists of three fundamental components: the dynamic collaborative network, the multi-view integration module, and the MIP adversarial supervision module.}
\vspace{-3mm}
\label{fig-all}
\end{figure*}
\subsection{Semi-Supervised Medical Segmentation}

Due to the scarcity of labeled data, semi-supervised learning has become increasingly prevalent in medical image segmentation, including the contrastive learning methods~\cite{SS-Net,SimCVD,RCPS}, pseudo-label methods~\cite{CoreNet,SSA-Net,COVID-19-seg,Liver-seg}, and consistency regularization methods~\cite{UA_MT,SASSNet,DTC,URPC,CPCL,DUW-SSL}.
The first two methods require intricate techniques to construct positive and negative samples and sophisticated strategies to refine pseudo-labels, respectively. The third method, consistency regularization, is gaining popularity due to its simplicity and effectiveness.

Consistency regularization methods are roughly categorized into  single model paradigm and mean teacher (MT) paradigm. 
The former optimize models by enforcing consistency between original and perturbed unlabeled samples \cite{CCT,ICT,DTC,DMT}.
Since a single model can only derive supervision from its own learning and training process, it struggles to overcome its inherent cognitive limitations, leading to significant cumulative errors.
In contrast, the latter imposes consistency constraints between the student and teacher models \cite{MT,Tug-MT,TCSE,Magicnet}. This paradigm introduces additional sources of supervision,  alleviating the  limitations of single model.

Despite their advantages, MT methods have drawbacks due to fixed teacher-student roles. Given the complexity of 3D vessel data, the teacher may not always outperform the student, leading to cognitive biases. To address this issue, this paper proposes a dynamic collaborative network that allows the two models to dynamically switch their teacher-student roles based on their timely performance.

\subsection{Vessel Segmentation}

Due to the complex nature of vessel appearance, numerous efforts have been made to achieve robust feature modeling. For example, DUNet~\cite{DUNet} incorporates deformable convolution~\cite{DCN} into a U-shaped network to adaptively adjust the receptive field according to the scale and shape of blood vessels. DSCnet~\cite{DSC} extends deformable convolution by proposing a dynamic snake convolution method based on topological geometric constraints, which accurately captures the characteristics of tubular structures.
MICNet~\cite{MIC-Net} employs dense hybrid dilated
convolution~\cite{DRN} in the connection layer, capturing richer contextual information while preserving feature resolution.

These methods typically design specialized layers tailored to the intricate tubular structure of blood vessels, resulting in enhanced adaptability to vascular images. However, such customized designs are often highly complex and lack generalizability. Additionally, these methods compute features globally and do not adequately emphasize essential local information. In contrast, this work introduces a straightforward multi-view integration module that effectively captures both the local vascular structure details and the global image context. This innovative approach requires only simple preprocessing of the input images, yet it achieves compelling vascular segmentation performance.


\section{Method}
\label{sec:method}

In this section, we introduce the DiCo in detail. It includes the dynamic collaboration network, multi-view integration module, and MIP adversarial supervision.

\subsection{Overview}
The overall framework is illustrated in Fig.~\ref{fig-all}.  Its core is the dynamic collaborative network, which comprises  a convolutional sub-network $\mathbf{M}_{1}$ and a transformer sub-network $\mathbf{M}_{2}$. 
In each training iteration, we use the better-performing sub-network as the teacher and the other as the student, establishing a dynamic supervision. 
Building on the dynamic collaborative network, we incorporate the multi-view integration module and the MIP projection adversarial module at the input and supervision stages, respectively, to improve the model's ability to handle complex vessel appearance. 

\subsection{Dynamic Collaborative  Network}

To overcome the cognitive bias of the static MT network, the dynamic collaborative network is proposed.
In this network, we utilize two sub-networks, $\mathbf{M}_{1}$ and $\mathbf{M}_{2}$, to dynamically supervise and enhance each other’s performance.
Specifically, in each training iteration, the labeled data $ \mathbf{X}^{l}$ and unlabeled  $\mathbf{X}^{u} $ are both fed into sub-networks $\mathbf{M}_{1}$ and $\mathbf{M}_{2}$ to generate the predictions, as summarized below:

\begin{equation}
\mathbf{Y}_{1}^{l},\mathbf{Y}_{1}^{u}=\mathbf{M}_{1}\left (\mathbf{X} ^{l}\right ),\mathbf{M}_{1}\left (\mathbf{X} ^{u} \right ),
\end{equation}
\begin{equation}
\mathbf{Y}_{2}^{l},\mathbf{Y}_{2}^{u}=\mathbf{M}_{2}\left (\mathbf{X} ^{l} \right ),\mathbf{M}_{2}\left (\mathbf{X} ^{u} \right ).
\end{equation}

Then, the predictions $\mathbf{Y}_{1}^{l}$ and $\mathbf{Y}_{2}^{l}$ for the labeled data are compared with the ground truth $\hat{\mathbf{Y}}^l$ to calculate the segmentation losses $\mathcal{L}_{1}^{l}$ and $\mathcal{L}_{2}^{l}$, respectively. By comparing $\mathcal{L}_{1}^{l}$ and $\mathcal{L}_{2}^{l}$, the teacher and student roles of $\mathbf{M}_{1}$ and $\mathbf{M}_{2}$ are determined.
Concretely, if $\mathcal{L}_{1}^{l} \leq \mathcal{L}_{2}^{l}$, $\mathbf{M}_{1}$ is considered superior to $\mathbf{M}_{2}$ for the current input data type, so we set $\mathbf{M}_{1}$ as the teacher model and $\mathbf{M}_{2}$ as the student model. In this case, the pseudo-label $\mathbf{\hat{Y}}^{u} = \mathbf{Y}_{1}^{u}$ and the supervised prediction $\mathbf{Y}_{o}^{u} = \mathbf{Y}_{2}^{u}$. Otherwise, the roles are reversed. It is formulated as follows:
\begin{equation}
\mathbf{\hat{Y}}^{u} = \begin{cases} 
\mathbf{Y}_{1}^{u} & \text{if } \mathcal{L}_{1}^{l} \leq \mathcal{L}_{2}^{l}, \\
\mathbf{Y}_{2}^{u} & \text{otherwise}.
\end{cases}
\end{equation}
\begin{equation}
\mathbf{Y}^{u}_{o} = \begin{cases} 
\mathbf{Y}_{2}^{u} & \text{if } \mathcal{L}_{1}^{l} \leq \mathcal{L}_{2}^{l}, \\
\mathbf{Y}_{1}^{u} & \text{otherwise}. 
\end{cases}
\end{equation}
The unsupervised loss $\mathcal{L}^{u}$ is then calculated by comparing $\mathbf{Y}_{o}^{u}$ and $\mathbf{\hat{Y}}^{u}$. This loss is used to update the current student model via gradient descent. In this way, our method enables dynamic teacher-student supervision between the two sub-networks, mitigating the cognitive bias problem inherent in conventional MT models.

To better leverage the advantages of this dynamic collaboration, we use different architectures for $\mathbf{M}_{1}$ and $\mathbf{M}_{2}$: $\mathbf{M}_{1}$ employs a convolutional architecture, while $\mathbf{M}_{2}$ uses a transformer architecture. This diversity allows the models to complement each other’s strengths and weaknesses.
$\mathbf{M}_{1}$'s convolutional architecture is adept at capturing fine-grained local features, while $\mathbf{M}_{2}$'s transformer architecture excels in understanding broader contextual relationships.
By supervising each other, $\mathbf{M}_{1}$ and $\mathbf{M}_{2}$ can dynamically refine their capabilities, leading to improved overall performance and a more nuanced understanding of the complex attributes of vessel data.

\subsection{Multi-view Integration Module}

\begin{figure}[t]
\centering
\includegraphics[width=1\columnwidth]{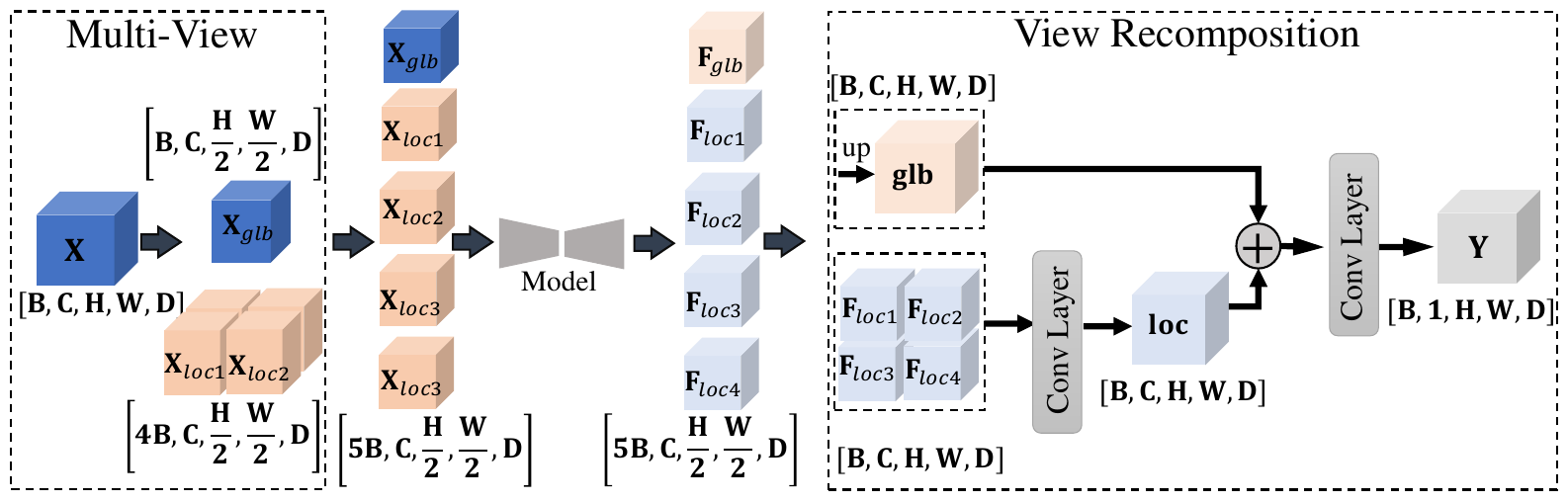} 
\vspace{-4mm}
\caption{Multi-view integration module.}
\label{MV}
\vspace{-4mm}
\end{figure}

We introduce a multi-view integration module that captures local perspectives of the input data, as illustrated in Fig.~\ref{MV}. This module highlights and integrates multiple local views of the original input to improve feature representation. Due to the Transformer's superior ability to handle unstructured information, we apply this module to the input of the Transformer sub-network $\mathbf{M}_{2}$. 

\textbf{Multi-view input.}
The original image $\mathbf{X} \in \mathbb{R}^{B \times C \times H \times W \times D}$  is first transformed into $\mathbf{X}_{loc} \in \mathbb{R}^{n_{1}n_{2}n_{3} \times B \times C \times H/n_{1} \times W/n_{2} \times D/n_{3}}$, where $n_{1}$$=$$2$, $n_{2}$$= $$2$, and $n_{3}$$=$$1$ in our implementation.
Subsequently, the original image is resized to $\mathbf{X}_{glb} \in \mathbb{R}^{B \times C \times H/n_{1} \times W/n_{2} \times D/n_{3}}$.
Next, $\mathbf{X}_{glb}$ and $\mathbf{X}_{loc}$ are concatenated along the B dimension to form the input $\mathbf{X}_{input} \in \mathbb{R}^{5B \times C \times H/2 \times W/2 \times D}$, incorporating both global and multiple local views. 
This input is then fed into the $\mathbf{M}_{2}$ model to generate a new feature with the same dimensions.
This multi-view mechanism enhances both global and local perspectives, facilitating a more comprehensive capture of the complex vessel appearance.

\textbf{View recomposition.}
After obtaining the output from network $\mathbf{M}_{2}$, the features from multiple local views need to be recomposed into a cohesive feature map to support the subsequent segmentation prediction. However, due to the feature extraction process, the spatial structure of the features may not be strictly preserved, potentially causing boundary misalignment.
To address this issue, we propose a view recomposition module to accurately reassemble the features. Specifically, the output features from model $\mathbf{M}_{1}$ are split along the batch dimension into five components: one global feature $\mathbf{F}_{\text{glb}}$ and four local features $\mathbf{F}_{\text{loc1}}, \mathbf{F}_{\text{loc2}}, \mathbf{F}_{\text{loc3}},$ and $\mathbf{F}_{\text{loc4}}$, each with dimensions $\mathbb{R}^{B \times C \times H/2 \times W/2 \times D}$.
The global feature $\mathbf{F}_{\text{glb}}$ is upsampled to the original dimension of $\mathbb{R}^{B \times C \times H \times W \times D}$.
For the local features, as illustrated in Fig.~\ref{MV}, we first recombine them into a feature map of size $\mathbb{R}^{B \times C \times H \times W \times D}$. Then, we apply 
two convolutional layers to smooth the boundaries of the recombined features.
Finally, the extracted features are combined with the global features, then passed through two convolutional layers to produce a refined mixed feature map that facilitates subsequent segmentation predictions.

\begin{figure}[t]
\centering 
\includegraphics[width=1\columnwidth]{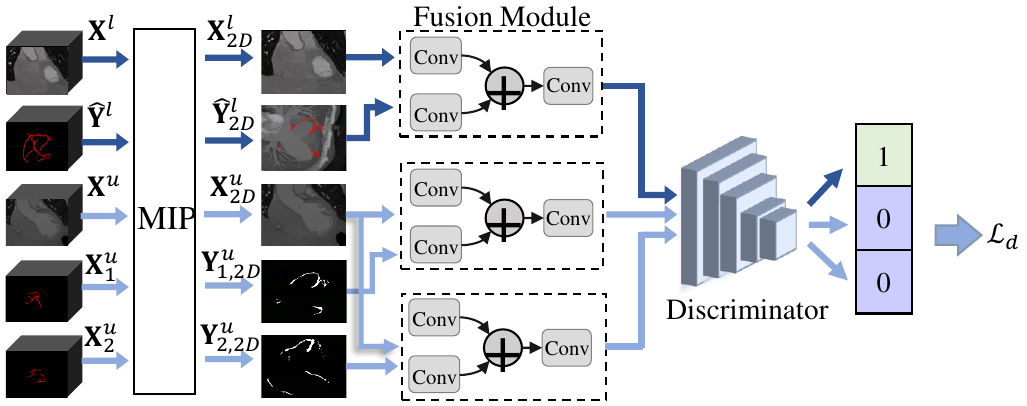} 
\vspace{-4mm}
\caption{MIP adversarial supervision module.}
\label{mip}
\vspace{-4mm}
\end{figure}

\begin{table*}[thb]
\renewcommand\arraystretch{0.9}
\caption{Comparison on ImageCAS dataset segmentation.}
\vspace{-3mm}
\label{tab:ImageCAS_result}
\centering

\resizebox{0.9\linewidth}{!}{
\begin{tabular}{cc|c|cc|ccc}
\noalign{\smallskip}
\toprule[2pt]
\noalign{\smallskip}
\multicolumn{2}{c|}{\multirow{2}{*}{Method}} &{\multirow{2}{*}{Source}}                      & \multicolumn{2}{c|}{Scans Used}                           & \multicolumn{3}{c}{Metrics}           \\ 
\cmidrule(r){4-5} \cmidrule(r){6-8}
\multicolumn{2}{c|}{}         & {} & \multicolumn{1}{c|}{\begin{tabular}[c]{@{}c@{}}L Volumes\end{tabular}} & U Volumes  & DSC$\uparrow$ (\%)      & NSD $\uparrow$(\%)   & ASD $\downarrow$(voxel)      \\ 
\noalign{\smallskip}
\hline
\noalign{\smallskip}
\multicolumn{1}{c|}{\multirow{9}{*}{Semi-supervised}} & {MT~\cite{MT}}&{ICLR'17}& \multicolumn{1}{c|}{45 } & 855          & 71.05        & 56.32         & 24.06                \\ 
\multicolumn{1}{c|}{} & {UA-MT~\cite{UA_MT}}&{MICCAI'19}& \multicolumn{1}{c|}{45 } & 855            & 70.11              & 55.83             & 23.11                           \\ 
\multicolumn{1}{c|}{} & {SASSNet~\cite{SASSNet}}&{MICCAI'20}& \multicolumn{1}{c|}{45 } & 855            & 72.73              & 58.54             &21.96                         \\ 
\multicolumn{1}{c|}{} & {SLCNet~\cite{SLCNet}}&{MICCAI'22}& \multicolumn{1}{c|}{45 } & 855            & 72.78               & 57.36            & 20.44                   \\ 
\multicolumn{1}{c|}{} & {MagicNet~\cite{Magicnet}}&{CVPR'23}& \multicolumn{1}{c|}{45 } & 855             & 71.88             & 57.50            & 22.10                   \\ 
\multicolumn{1}{c|}{} & {CAML~\cite{CAML}}&{MICCAI'23}& \multicolumn{1}{c|}{45 } & 855             & 71.66            & 58.03           & 23.73                  \\ 
\multicolumn{1}{c|}{} & {CauSSL~\cite{Caussl}}&{ICCV'23}& \multicolumn{1}{c|}{45 } & 855             & 69.14           & 53.25          & 24.79                  \\ 
\multicolumn{1}{c|}{} & {GuidedNet~\cite{GuidedNet}}&{ACMMM'24}& \multicolumn{1}{c|}{45 } & 855            & 65.24              & 50.63        & 30.21                   \\ 
\multicolumn{1}{c|}{} & {\textbf{DiCo}}&{\textbf{Ours}}& \multicolumn{1}{c|}{45 } & 855             & \textbf{73.79}              & \textbf{58.59}        & \textbf{20.00}                   \\

\noalign{\smallskip}
\hline
\noalign{\smallskip}

\multicolumn{1}{c|}{\multirow{2}{*}{Full-supervised}} & {VNet~\cite{VNet}}&{ICCV'16}& \multicolumn{1}{c|}{900 } & 0           & 71.19         & 55.36         & 23.88                 \\ 
\multicolumn{1}{c|}{} &{CTNet~\cite{CTN}}&{ BIBM'22}& \multicolumn{1}{c|}{900} & 0           & 79.71        & 69.31         & 11.97              \\

\multicolumn{1}{c|}{} & {ERNet~\cite{ERNet}}&{Med Image Anal '22}& \multicolumn{1}{c|}{900} & 0           & 76.25        & 64.56       & 18.92               \\ 

\multicolumn{1}{c|}{} & {DSCNet~\cite{DSCNet}}&{ ICCV'23}& \multicolumn{1}{c|}{900} & 0           & 73.49          & 58.06       & 22.78                \\ 

\noalign{\smallskip}
\bottomrule[2pt]
\noalign{\smallskip}
\end{tabular}}
\end{table*}

\begin{figure*}[t]
\vspace{-2mm}
\centering
\includegraphics[width=0.95\textwidth]{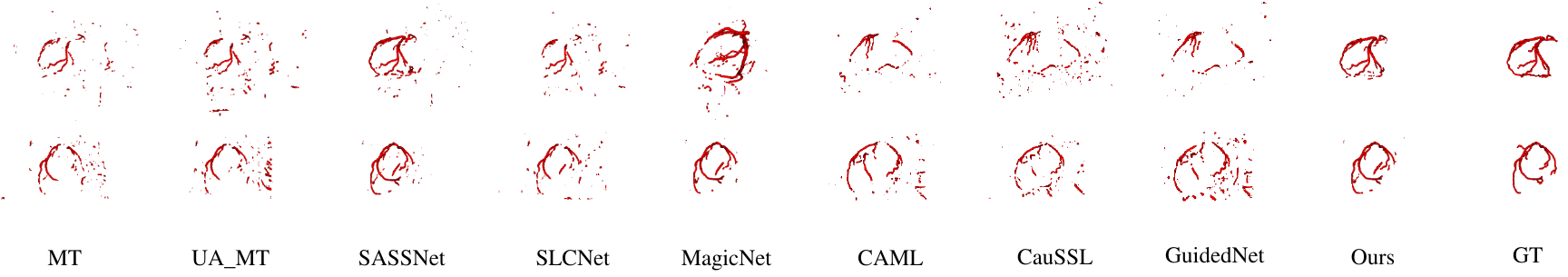} 
\vspace{-2mm}
\caption{Visual segmentation examples from ImageCAS dataset. }
\label{imageCAS-vision}
\vspace{-3mm}
\end{figure*}

\begin{table*}[thb]
\renewcommand\arraystretch{0.9}
\caption{Comparison on CAS2023 dataset segmentation.}
\vspace{-3mm}
\label{tab:cAS2023_result}
\centering

\resizebox{0.9\linewidth}{!}{
\begin{tabular}{cc|c|cc|ccc}
\noalign{\smallskip}
\toprule[2pt]
\noalign{\smallskip}
\multicolumn{2}{c|}{\multirow{2}{*}{Method}} &{\multirow{2}{*}{Source}}                      & \multicolumn{2}{c|}{Scans Used}                           & \multicolumn{3}{c}{Metrics}           \\ 
\cmidrule(r){4-5} \cmidrule(r){6-8}
\multicolumn{2}{c|}{}         & {} & \multicolumn{1}{c|}{\begin{tabular}[c]{@{}c@{}}L Volumes\end{tabular}} & U Volumes  & DSC$\uparrow$ (\%)      & NSD $\uparrow$(\%)   & ASD $\downarrow$(voxel)      \\ 
\noalign{\smallskip}
\hline
\noalign{\smallskip}
\multicolumn{1}{c|}{\multirow{9}{*}{Semi-supervised}} &  {MT~\cite{MT}}&{ICLR'17}& \multicolumn{1}{c|}{5 } & 85         & 73.94        & 58.33        & 15.21               \\ 
\multicolumn{1}{c|}{} & {UA-MT~\cite{UA_MT}}&{MICCAI'19}& \multicolumn{1}{c|}{5 } & 85            & 77.31             & 59.22           & 5.89                           \\ 
\multicolumn{1}{c|}{} & {SASSNet~\cite{SASSNet}}&{MICCAI'20}& \multicolumn{1}{c|}{5 } & 85            & 75.43              & 58.82             &7.67                         \\ 
\multicolumn{1}{c|}{} & {SLCNet~\cite{SLCNet}}&{MICCAI'22}& \multicolumn{1}{c|}{5 } & 85            & 77.21               & 61.65           & 3.65                   \\ 
\multicolumn{1}{c|}{} & {MagicNet~\cite{Magicnet}}&{CVPR'23}& \multicolumn{1}{c|}{5 } & 85            & 84.13             & 72.73           & 1.92                   \\ 
\multicolumn{1}{c|}{} & {CAML~\cite{CAML}}&{MICCAI'23}& \multicolumn{1}{c|}{5 } & 85             & 79.64            & 67.43           & 2.19                  \\ 
\multicolumn{1}{c|}{} & {CauSSL~\cite{Caussl}}&{ICCV'23}& \multicolumn{1}{c|}{5 } & 85             & 76.72           & 62.48          & 2.22                  \\ 
\multicolumn{1}{c|}{} & {GuidedNet~\cite{GuidedNet}}&{ACMMM'24}& \multicolumn{1}{c|}{5 } & 85            & 81.60              & 69.00        & 2.01                  \\ 
\multicolumn{1}{c|}{} & {\textbf{DiCo}}&{\textbf{Ours}}& \multicolumn{1}{c|}{5 } & 85            & \textbf{86.05}              & \textbf{74.35}        & \textbf{1.49}                   \\

\noalign{\smallskip}
\hline
\noalign{\smallskip}
\multicolumn{1}{c|}{\multirow{2}{*}{Full-supervised}} & {VNet~\cite{VNet}}&{ICCV'16}& \multicolumn{1}{c|}{90 } & 0           & 67.25         & 43.38        & 21.36                 \\ 
\multicolumn{1}{c|}{} &{CTNet~\cite{CTN}}&{ BIBM'22}& \multicolumn{1}{c|}{90} & 0           & 77.51        & 63.24        & 8.76             \\

\multicolumn{1}{c|}{} & {ERNet~\cite{ERNet}}&{Med Image Anal '22}& \multicolumn{1}{c|}{90} & 0           & 74.62        &59.78       & 12.43               \\ 

\multicolumn{1}{c|}{} & {DSCNet~\cite{DSCNet}}&{ ICCV'23}& \multicolumn{1}{c|}{90} & 0         & 83.14         & 70.09       & 2.69                \\ 

\noalign{\smallskip}
\bottomrule[2pt]
\noalign{\smallskip}
\end{tabular}}
\end{table*}

\begin{figure*}[t]
\vspace{-2mm}
\centering
\includegraphics[width=0.95\textwidth]{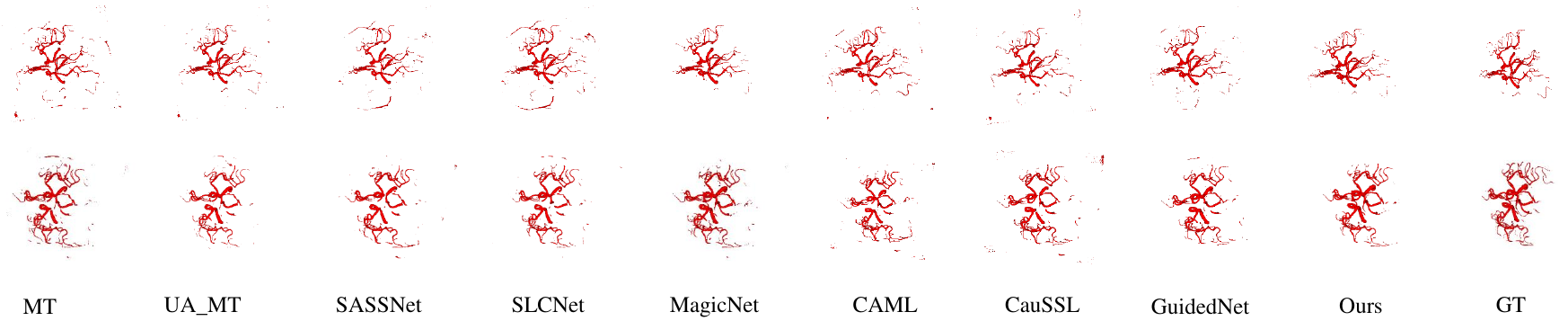} 
\vspace{-2mm}
\caption{Visual segmentation examples from CAS2023 dataset.}
\vspace{-3mm}
\label{CAS2023-vision}
\end{figure*}

\begin{table*}[thb]
\renewcommand\arraystretch{0.9}
\caption{Comparison on Parse2022 dataset segmentation.}
\vspace{-3mm}
\label{tab:parse2022_result}
\centering

\resizebox{0.9\linewidth}{!}{
\begin{tabular}{cc|c|cc|ccc}
\noalign{\smallskip}
\toprule[2pt]
\noalign{\smallskip}
\multicolumn{2}{c|}{\multirow{2}{*}{Method}} &{\multirow{2}{*}{Source}}                      & \multicolumn{2}{c|}{Scans Used}                           & \multicolumn{3}{c}{Metrics}           \\ 
\cmidrule(r){4-5} \cmidrule(r){6-8}
\multicolumn{2}{c|}{}         & {} & \multicolumn{1}{c|}{\begin{tabular}[c]{@{}c@{}}L Volumes\end{tabular}} & U Volumes  & DSC$\uparrow$ (\%)      & NSD $\uparrow$(\%)   & ASD $\downarrow$(voxel)      \\ 
\noalign{\smallskip}
\hline
\noalign{\smallskip}
\multicolumn{1}{c|}{\multirow{9}{*}{Semi-supervised}} &  {MT~\cite{MT}}&{ICLR'17}& \multicolumn{1}{c|}{5 } & 85         & 58.36        & 42.58        & 11.37               \\ 
\multicolumn{1}{c|}{} & {UA-MT~\cite{UA_MT}}&{MICCAI'19}& \multicolumn{1}{c|}{5 } & 85            &62.70             & 45.72           & 10.51                           \\ 
\multicolumn{1}{c|}{} & {SASSNet~\cite{SASSNet}}&{MICCAI'20}& \multicolumn{1}{c|}{5 } & 85            & 68.33              & 29.19             &7.57                         \\ 
\multicolumn{1}{c|}{} & {SLCNet~\cite{SLCNet}}&{MICCAI'22}& \multicolumn{1}{c|}{5 } & 85            & 66.13               & 48.87           & 8.18                   \\ 
\multicolumn{1}{c|}{} & {MagicNet~\cite{Magicnet}}&{CVPR'23}& \multicolumn{1}{c|}{5 } & 85            & 69.19             & 53.25           & \textbf{5.53}                   \\ 
\multicolumn{1}{c|}{} & {CAML~\cite{CAML}}&{MICCAI'23}& \multicolumn{1}{c|}{5 } & 85             & 66.75           & 50.22          & 7.27                     \\ 
\multicolumn{1}{c|}{} & {CauSSL~\cite{Caussl}}&{ICCV'23}& \multicolumn{1}{c|}{5 } & 85             & 66.45              & 48.88        & 7.94                      \\ 
\multicolumn{1}{c|}{} & {GuidedNet~\cite{GuidedNet}}&{ACMMM'24}& \multicolumn{1}{c|}{5 } & 85         &   68.80              & 51.80       & 6.74              \\ 
\multicolumn{1}{c|}{} & {\textbf{DiCo}}&{\textbf{Ours}}& \multicolumn{1}{c|}{5 } & 85            & \textbf{70.93}              & \textbf{55.26}       & 5.74                   \\

\noalign{\smallskip}
\hline
\noalign{\smallskip}
\multicolumn{1}{c|}{\multirow{2}{*}{Full-supervised}} & {VNet~\cite{VNet}}&{ICCV'16}& \multicolumn{1}{c|}{90 } & 0           & 65.53         & 48.87        & 8.08                 \\ 
\multicolumn{1}{c|}{} & {CTNet~\cite{CTN}}&{ BIBM'22}& \multicolumn{1}{c|}{90} & 0           &73.12        & 58.92         & 5.88               \\

\multicolumn{1}{c|}{} & {ERNet~\cite{ERNet}}&{Med Image Anal '22}& \multicolumn{1}{c|}{90} & 0           & 76.39        & 64.72       & 3.81              \\ 
\multicolumn{1}{c|}{} & {DSCNet~\cite{DSCNet}}&{ ICCV'23}& \multicolumn{1}{c|}{90} & 0         & 75.04         & 59.66       & 4.49                \\ 
\noalign{\smallskip}
\bottomrule[2pt]
\noalign{\smallskip}
\end{tabular}}
\end{table*}

\begin{figure*}[t]
\vspace{-2mm}
\centering
\includegraphics[width=0.95\textwidth]{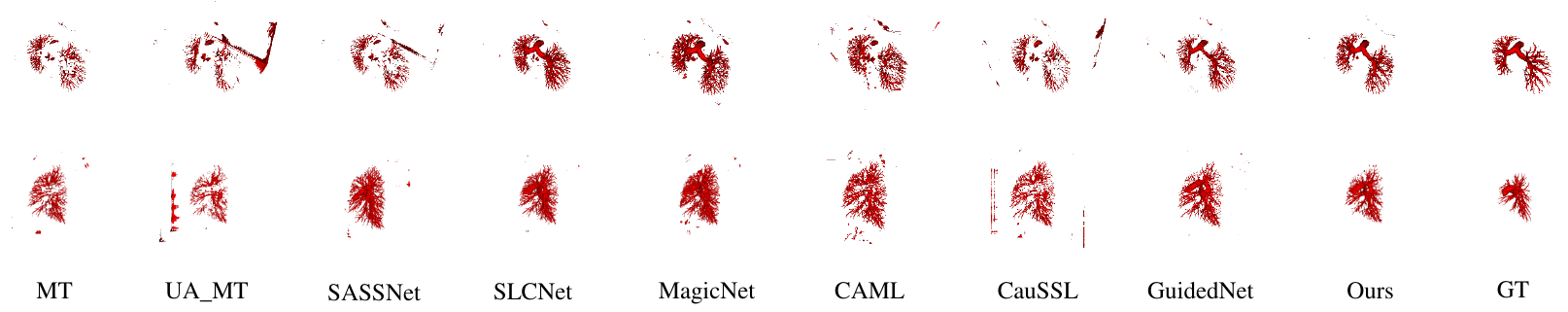} 
\vspace{-2mm}
\caption{Visual segmentation examples from Parse2022 dataset.}
\vspace{-3mm}
\label{parse2022-vision}
\end{figure*}

\subsection{MIP Adversarial Supervision}

3D vessels typically exhibit complex, elongated tubular structures. To enforce constraints based on vessel shape priors, we propose a maximum-intensity-projection (MIP) based adversarial supervision, as illustrated in Fig.~\ref{mip}. In this approach, a discriminator is trained to evaluate whether the unlabeled predicted masks capture a similar shape style to the labeled ground-truth masks. 
We then train networks $\mathbf{M}1$ and $\mathbf{M}2$ to generate masks that deceive the discriminator, thereby improving their ability to produce well-shaped vessel masks. 
However, using precise 3D data can pose a risk of overfitting. To mitigate this, we project the 3D data into 2D before applying the adversarial supervision.

Specifically, we use MIP to project the labeled image \( \mathbf{X}^l \) with its corresponding label \( \mathbf{\hat{Y}}^l \), as well as unlabeled image \( \mathbf{X}^u \) with its predictions \( \mathbf{Y}^u_1 \) and \( \mathbf{Y}^u_2 \), into 2D along the depth dimension.  Mathematically, for the 3D images  \( \mathbf{X}^l \) and \( \mathbf{X}^u \), the MIP along the depth dimension \( z \) is computed as follows:
\begin{equation}
\mathbf{X}^l_{2D}(x, y) = \max_{z} \mathbf{X}^l(x, y, z),
\end{equation}
\begin{equation}
\mathbf{X}^u_{2D}(x, y) = \max_{z} \mathbf{X}^u(x, y, z),
\end{equation}
where \( \mathbf{X}^l_{2D} \) and \( \mathbf{X}^u_{2D} \) denote the projected labeled and unlabeled images, respectively. Here, \( (x, y) \) represent the spatial coordinates in the 2D projection, and \( z \) denotes the depth dimension. Similarly, for the label \( \hat{Y} \) and the predictions \( \mathbf{Y}^u_1 \) and \( \mathbf{Y}^u_2 \), the 2D projection using MIP is given by:
\begin{equation}
\mathbf{\hat{Y}}^l_{2D}(x, y) = \max_{z} \mathbf{\hat{Y}}^l(x, y, z),
\end{equation}
\begin{equation}
\mathbf{Y}^u_{1,2D}(x, y) = \max_{z} \mathbf{Y}^u_{1}(x, y, z),
\end{equation}
\begin{equation}
\mathbf{Y}^u_{2,2D}(x, y) = \max_{z} \mathbf{Y}^u_{2}(x, y, z).
\end{equation}
Here, \( \mathbf{Y}^u_{1,2D} \) and \( \mathbf{Y}^u_{2,2D} \) denotes the projected predictions for the unlabeled image, while \( \mathbf{\hat{Y}}^l_{2D} \) represents the projected label of the labeled image. We then fuse these labels or predictions with their corresponding images in a structured manner, as summarized:
\begin{equation}
\mathbf{\hat{O}}^l_{2D} = \mathbf{F}(\mathbf{\hat{Y}}^l_{2D}, \mathbf{X}^l_{2D}),
\end{equation}
\begin{equation}
\mathbf{O}^u_{1,2D} = \mathbf{F}(\mathbf{Y}^u_{1,2D}, \mathbf{X}^u_{2D}),
\end{equation}
\begin{equation}
\mathbf{O}^u_{2,2D} = \mathbf{F}(\mathbf{Y}^u_{2,2D}, \mathbf{X}^u_{2D}).
\end{equation}
Here, \( \mathbf{F}(\cdot) \) denotes a fusion module, as illustrated by the dotted box in Fig.~\ref{mip}. The fused outputs are represented as $\mathbf{\hat{O}}^l_{2D}$, $\mathbf{O}^u_{1,2D}$, and $\mathbf{O}^u_{2,2D}$. 
We then feed the three fused images into a discriminator \( \mathbf{D}(\cdot) \), training it to distinguish between images generated using the ground-truth mask and those generated using predicted masks. The loss function is the binary cross-entropy (BCE) loss. In this setup, the labels for $\mathbf{O}^u_{1,2D}$ and $\mathbf{O}^u_{2,2D}$ are set to 0, while the label for $\mathbf{\hat{O}}^l_{2D}$ is set to 1. This is summarized as:
\begin{equation}
\begin{aligned}
\mathcal{L}_{d} = & \mathcal{L}_{bce}(\mathbf{D}(\hat{O}^l_{2D}),1) + \mathcal{L}_{bce}(\mathbf{D}(\mathbf{O}^u_{1,2D}),0) \\
& + \mathcal{L}_{bce}(\mathbf{D}(\mathbf{O}^u_{2,2D}),0).
\end{aligned}
\end{equation}
Then, we train $\mathbf{M}_1$ and $\mathbf{M}_2$ to deceive the discriminator $\mathbf{D}$ by minimizing an adversarial loss $\mathcal{L}_{adv}$:
\begin{equation}
\begin{aligned}
\mathcal{L}_{adv} =
\mathcal{L}_{bce}(\mathbf{D}(\mathbf{O}^u_{1,2D}),1) + \mathcal{L}_{bce}(\mathbf{D}(\mathbf{O}^u_{2,2D}),1).
\end{aligned}
\end{equation}

\subsection{Loss Functions}
To supervise the segmentation of models $\mathbf{M}1$ and $\mathbf{M}2$, 
we employ a combination of Dice loss and cross-entropy loss:
\begin{equation}
\mathcal{L}_{seg}=\alpha\mathcal{L}_{dice}+\beta\mathcal{L}_{ce},
\end{equation}
\begin{equation}
\mathcal{L}_{1}^{l}=\mathcal{L}_{seg}(\mathbf{Y}_{1}^{l},\mathbf{\hat{Y}}^{l}),
\end{equation}
\begin{equation}
\mathcal{L}_{2}^{l}=\mathcal{L}_{seg}(\mathbf{Y}_{2}^{l},\mathbf{\hat{Y}}^{l}),
\end{equation}
where $\alpha$ and $\beta$ are regularization parameters. 
For the dynamic collaborative network, the unsupervised loss function is as follows:
\begin{equation}
\mathcal{L}^{u} = \mathcal{L}_{seg}(\mathbf{Y}^{u}_{o}, \mathbf{\hat{Y}}^{u}).
\end{equation}
The total loss function is given by:
\begin{equation}
\mathcal{L}=\mathcal{L}_{1}^{l} + \mathcal{L}_{2}^{l} + \mathcal{L}^{u} + \mathcal{L}_{adv}.
\end{equation}

\section{Experiments}
\label{sec:exp}

\subsection{Datasets and Metrics }
We use three vessel datasets for training and evaluation, 5\% of the training data are selected as labeled volumes, while the remaining data are treated as unlabeled volumes during the training process:
\begin{itemize}
\item
\textbf{ImageCAS} 
\cite{ImageCAS} comprises 1,000 3D coronary vessel computed tomography angiography (CTA) images, with voxel dimensions ranging from $512 \times 512 \times (206 - 275)$. Of these, 900 images are allocated for training, and 100 for evaluation.
\item 
~\textbf{CAS2023} 
\cite{cas2023} is a cerebral artery vessel segmentation dataset from the MICCAI 2023 challenge, containing 100 public magnetic resonance angiography (MRA) samples. This dataset includes 90 images for training and 10 for evaluation. 
\item 
~\textbf{Parse2022} \cite{parse2022} provides computed tomography (CT) images for pulmonary artery vessel segmentation from the MICCAI 2022 challenge.This dataset includes 90 images for training and 10 for evaluation. 
\end{itemize}

The performance is assessed by using three  metrics: the Dice similarity coefficient (DSC) for region sensitivity, the normalized surface Dice coefficient (NSD) for evaluating surface overlap precision, and the average surface distance (ASD) for edge sensitivity. The DSC is generally regarded as a primary metric for medical image segmentation.

\subsection{Implementation Details}
All experiments in this study are conducted using Python 3.9 and PyTorch 2.2, with training and evaluation carried out on an NVIDIA 3090 GPU. 

For the model $\mathbf{M}1$, we employ VNet~\cite{VNet}, a widely recognized convolutional network for medical image segmentation. For $\mathbf{M}2$, we select UNETR~\cite{UNETR}, a well-established vision transformer network. Details are available in \textit{supplementary materials}. During inference, the final prediction is obtained from the VNet output, consistent with the approach used in other methods that utilize VNet. 

Our model is trained using the AdamW optimizer for 40,000 iterations, with an initial learning rate $\text{lr}_{\text{base}}$ of $1$$e$$-2$.
A learning rate decay strategy is employed, defined by:
 \begin{equation}
\text{lr} = \text{lr}_{\text{base}} \times \left( 1 - \frac{t}{T} \right)^{\gamma}.
\end{equation}
Here, $t$ represents the current training iteration, $T$ denotes the total number of iterations, $\gamma$ is the exponent used to adjust the rate of decay.
The batch size is set to $2$. We employ a center-crop with size $96$$\times$$96$$\times$$96$ for the input volumes. During the inference phase, a sliding window approach is employed to generate the final result for the entire volume.

\subsection{Comparisons with State-of-the-arts}
We compare our DiCo method with state-of-the-art (SOTA) self-supervised medical image segmentation methods, including MT~\cite{MT}, UA-MT~\cite{UA_MT}, SASSNet~\cite{SASSNet}, SLCNet~\cite{SLCNet}, MagicNet~\cite{Magicnet}, CAML~\cite{CAML}, CauSSL~\cite{Caussl}, and GuidedNet~\cite{GuidedNet}, as well as fully-supervised methods designed for vessel data, such as CTNet~\cite{CTN}, DSCNet~\cite{DSCNet}, and ERNet~\cite{ERNet}. We also incorporate the baseline method VNet~\cite{VNet} in the comparison, which is a fully-supervised medical image segmentation method.

 \textbf{ImageCAS.} As shown in Tab~\ref{tab:ImageCAS_result}, on the ImageCAS dataset, our DiCo method achieves a DSC of 73.79\%, an NSD of 58.59\%, and an ASD of 20.00 voxels. These results surpass other semi-supervised methods, demonstrating the superiority of the DiCo method. Moreover, DiCo achieves comparable performance to fully supervised methods while using only 5\% of the labeled data, underscoring its efficiency in leveraging limited annotations. Visualization results are provided in Fig.~\ref{imageCAS-vision}.
 

\textbf{CAS2023.} As shown in Tab~\ref{tab:cAS2023_result}, on the CAS2023 dataset, our DiCo method achieves the top performance with 86.05\% in DSC, 74.35\% in NSD, and 1.49 voxels in ASD, outperforming all compared semi-supervised methods across each metric. In particular, DiCo surpasses the second-best MagicNet by 2.28\% in the primary metric DSC. 
Furthermore, DiCo demonstrates comparable performance to fully supervised methods that rely on significantly more labeled data, underscoring DiCo’s efficiency in semi-supervised learning. Examples of segmentation results are shown in Fig.~\ref{CAS2023-vision}.

\begin{table}[t]
\centering
\setlength{\tabcolsep}{3mm}{
\small
\scalebox{0.9}{
\begin{tabular}{c c| c c c c }
\toprule[2pt]
Dataset & Metric & ~C+T~ & ~C+C~ & ~T+T~ & ~MT~  \\
\midrule 

\multirow{3}{*}{ \emph{ImageCAS}} 
    &\emph{DSC$\uparrow$}&70.37 &69.45&62.67&\textbf{71.05} \\ 
    & \emph{NSD$\uparrow$} & \textbf{56.36} & 53.42 & 47.54 & 56.32 \\ 
    & \emph{ASD$\downarrow$}  &\textbf{23.87}& 25.89 & 32.47 &  24.06 \\
 \hline

\multirow{3}{*}{ \emph{CAS2023}} 
    &\emph{DSC$\uparrow$}&\textbf{83.59} &82.74&78.56&73.94 \\ 
    & \emph{NSD$\uparrow$} & \textbf{73.44} & 70.89 & 67.28 & 58.33 \\ 
    & \emph{ASD$\downarrow$}  & \textbf{1.79}& 1.97 & 2.17 & 15.21 \\
 \hline
\multirow{3}{*}{\emph{Parse2022}}    
& \emph{DSC$\uparrow$} & \textbf{63.85}   &  62.77   & 62.89 &  58.36 \\
& \emph{NSD$\uparrow$} & \textbf{43.39}   &  43.21   & 42.45 &  42.58  \\
& \emph{ASD$\downarrow$} & \textbf{9.02}  &  10.84   & 9.65 &  11.37 \\
\bottomrule[2pt]
\end{tabular}}
}
\vspace{-1mm}
\caption{Ablation study results of the DiCo architecture. MT stands for the basic mean teacher architecture. C represents the CNN network V-Net, and T denotes the ViT network UNETR. C+C, C+T, and T+T represent DiCo with different combinations of sub-networks.}
\vspace{-3mm}
\label{table:ablation1}
\end{table}

\textbf{Parse2022.} As shown in Tab~\ref{tab:parse2022_result}, on the Parse2022 dataset, our DiCo method achieves a DSC of 70.93\%, an NSD of 55.26\%, and an ASD of 5.74 voxels. Both the DSC and NSD are superior to other semi-supervised methods. Specifically, DiCo surpasses the recently proposed GuidedNet by 3.10\% in DSC and by 6.68\% in NSD. In terms of the ASD metric, DiCo also delivers a competitive result, trailing MagicNet by only 0.21 voxels while surpassing all other semi-supervised  methods. Representative segmentation results are shown in Fig.~\ref{parse2022-vision}.






\begin{table}[t]
\centering
\setlength{\tabcolsep}{1.5mm}{
\small
\scalebox{0.9}{
\begin{tabular}{c c| c c c c c }
\toprule[2pt]
Dataset & Metric & ~MT~ & ~Base~ & ~+MIP~ & ~+MV~ & ~All~ \\
\midrule 

\multirow{3}{*}{\emph{ImageCAS}} 
& \emph{DSC$\uparrow$}  &  71.05  &  70.37  &  71.15  &  72.37  &  \textbf{73.79}  \\
& \emph{NSD$\uparrow$}  &  56.32  &  56.36  &  53.64  &  57.36  &  \textbf{58.59}  \\
& \emph{ASD$\downarrow$}  &  24.06  &  23.87  &  21.97  &  \textbf{19.44}  &  20.00  \\
\hline

\multirow{3}{*}{\emph{CAS2023}}   
& \emph{DSC$\uparrow$} &   73.94  &  83.59  &   84.42 & 85.63  &  \textbf{86.05}  \\
& \emph{NSD$\uparrow$} &   58.33  &  73.44  &   73.24 & 73.40  &  \textbf{74.35}  \\
& \emph{ASD$\downarrow$} &   15.21  &  1.79  &   2.31 & 1.63  &  \textbf{1.49}  \\
\hline
\multirow{3}{*}{\emph{Parse2022}}    
& \emph{DSC$\uparrow$} &  58.36       &  63.85  &  66.29 & 68.46 &  \textbf{70.93}  \\
& \emph{NSD$\uparrow$} &  42.58       &  47.39  &  50.15 & 52.14 &  \textbf{55.26}  \\
& \emph{ASD$\downarrow$} &  11.37     &  9.02  &  7.24 & 5.92 &  \textbf{5.74}  \\
\bottomrule[2pt]
\end{tabular}}
}
\vspace{-2mm}
\caption{Component-wise study. MT represents the basic mean teacher architecture,  MIP denotes our MIP adversarial supervision, MV indicates our multi-view integration model, Base denotes our DiCo without MIP and MV, and All is our default DiCo model.}
\vspace{-1mm}
\label{table:ablation2}
\end{table}

\begin{table}[t]
\centering
\resizebox{\linewidth}{!}{
\setlength{\tabcolsep}{3.5mm}{
\begin{tabular}{c c|c c c }
\toprule[2pt]
\multirow{2}{*}{Datasets} & \multirow{2}{*}{Method} & \multicolumn{3}{c}{Metric} \\ \cline{3-5}
  &   
  & \emph{DSC$\uparrow$}  
  & \emph{NSD$\uparrow$} & \emph{ASD$\downarrow$}\\
  \midrule[0.5pt]
\multirow{2}{*}{\emph{ImageCAS}}
& 3D   &  60.79    &   43.73  &  46.31    \\ 
& 2D  &  \textbf{71.15}    & \textbf{53.64}  & \textbf{21.97}        \\         \hline
\multirow{2}{*}{\emph{CAS2023}}
& 3D    &  83.25   &   71.39  &   2.49   \\ 
& 2D   &  \textbf{84.42}    & \textbf{73.24}  &   \textbf{2.31}      \\         \hline
\multirow{2}{*}{\emph{Pares2022}}
& 3D     &  59.33   &  46.92   &  12.68    \\ 
& 2D    & \textbf{66.29}     & \textbf{50.15}  &   \textbf{7.24}      \\   
\bottomrule[2pt]
\end{tabular}
}}
\vspace{-2mm}
\caption{Ablation study of the 2D projection in MIP adversarial supervision.}
\vspace{-3mm}
\label{table:ablation4}
\end{table}

\subsection{Ablation Study}
\textbf{DiCo variants.} 
We conducted experiments to explore various  combinations of sub-networks, including CNN+ViT (our default configuration, denoted as C+T), CNN+CNN (denoted as C+C), and ViT+ViT (denoted as T+T). 
The results are reported in Tab.~\ref{table:ablation1}. Among these configurations, the C+T combination achieves the best performance. 
This superior performance can be attributed to the complementary strengths of the CNN and ViT: the CNN effectively captures fine-grained local features, while the transformer excels at understanding broader contextual relationships. By supervising each other, the two sub-networks dynamically enhance their capabilities, leading to a more nuanced understanding of the complex characteristics of vessel data.

\textbf{DiCo \emph{v.s.} MT.} 
Tab.~\ref{table:ablation1} also presents the results of the mean teacher (MT) method, which employs static supervision. Our DiCo approach demonstrates strong overall effectiveness, highlighting the power of dynamic collaboration. 
Notably, in the DiCo approach, our default C+T configuration surpasses the aligned MT method by an average of 7.16\% in DSC across three vessel datasets.

\textbf{Component-wise Study.}
We conduct experiments to explore the impact of each component in DiCo, with the results presented in Tab.~\ref{table:ablation2}. In this table, MT represents the basic mean teacher architecture, MIP denotes our MIP adversarial supervision, MV indicates our multi-view integration model, Base denotes our DiCo without MIP and MV, and All is our default DiCo model. 

Compared to the MT method, even our base DiCo model demonstrates a significant improvement, with a 13.05\% increase on the CAS2023 dataset. Building on this, the MIP adversarial supervision approach further enhances performance across the three datasets by 1.11\%, 1.00\%, and 3.82\% in DSC score, respectively, showcasing its effectiveness. The MV model improves performance by 2.84\%, 2.44\%, and 7.22\% on the three datasets, underscoring the benefits of integrating multiple data views. Ultimately, the combination of all three designs delivers the best performance across all datasets.

\textbf{3D v.s. 2D adversarial supervision.}
We conduct experiments to assess the impact of using MIP to project 3D data into 2D for adversarial supervision.
The results presented in Tab.~\ref{table:ablation4} indicate that projecting the 3D data to 2D improves the DSC scores by 17.04\%, 1.41\%, and 11.73\% points across the three datasets, respectively. These findings demonstrate the effectiveness of our 2D projection approach for the adversarial supervision.



\textbf{Impact of loss function.}
We investigate the impact of the loss functions used, with the results detailed in Table~\ref{tab-loss}. The evaluated dataset is CAS2023. As shown in Rows \#1 and \#2, relying solely on cross-entropy loss $\mathcal{L}{ce}$ or Dice loss $\mathcal{L}{dice}$ can result in performance degradation. Combining these two loss functions, as demonstrated in Row \#3, leads to improved performance. Row \#4 demonstrates that incorporating our unsupervised loss \(\mathcal{L}_{u}\) significantly enhances performance. Row\#5 shows that our MIP adversarial supervision further improves the performance.

\begin{table}[t]
\centering
\setlength{\tabcolsep}{1.5mm}{
\small
\scalebox{0.95}{
\begin{tabular}{c|l| c c c}
\toprule[2pt]
\# &Loss  & DSC$\uparrow$ & NSD$\uparrow$ &ASD$\downarrow$ \\
\midrule 
1&$ \mathcal{L}_{ce} $~~ &   74.23  & 59.56  &  11.47  \\
2&$ \mathcal{L}_{dice} $        &  76.00   & 62.25  & 8.90  \\
3&$ \mathcal{L}_{seg} $        &  78.41  &  63.69  &  5.72  \\
4&$ \mathcal{L}_{seg}+\mathcal{L}_{u} $    &  85.63   &73.40  & 1.63  \\
5&$ \mathcal{L}_{seg}+\mathcal{L}_{u}+\mathcal{L}_{adv} $  & \textbf{86.05}   &\textbf{74.35}   & \textbf{1.49} \\
\bottomrule[2pt]
\end{tabular}}
}
\vspace{-2mm}
\caption{The impact of the loss function is verified on the CAS2023 dataset.}
\vspace{-3mm}
\label{tab-loss}
\end{table}

\section{Conclusion}
\label{sec:con}
This work introduces DiCo, a new semi-supervised 3D vessel segmentation framework that enables dynamic collaboration between two models. By dynamically alternating teacher and student roles based on model performance, DiCo mitigates the cognitive biases inherent in conventional static semi-supervision approaches. 
Additionally, the integration of a multi-view module and MIP-based adversarial supervision further enhances segmentation quality. Experimental results demonstrate that DiCo is effective, achieving competitive performance compared to state-of-the-art medical segmentation methods. We hope this work provides a new perspective on learning for 3D vessel segmentation.

\noindent\textbf{Acknowledgements.}
This work was supported by the National Natural Science Foundation of China under Grant 62431004 and 62276046, and by Dalian Science and Technology Innovation Foundation under Grant 2023JJ12GX015.

{
    \small
    \bibliographystyle{ieeenat_fullname}
    \bibliography{main}
}


\end{document}


\clearpage
\setcounter{page}{1}
\maketitlesupplementary

\section{Rationale}
\label{sec:rationale}
In this appendix, we provide additional content to complement the main manuscript:
\begin{itemize}[leftmargin=0.468cm]
    \item{More Details About Datasets}
    \item{Relative \emph{vs.} Absolute Evaluation Analysis}
    \item{Additional Experimental Results}
    \item{Evaluation Prompts}
\end{itemize}

\appendix
\section{More Details About Datasets}
This section provides additional details about the two types of datasets utilized in our experiments. The first includes a range of publicly available medical text and image-text datasets, with detailed descriptions of their types, modalities, and sizes. The second is our self-constructed MediScope dataset, which showcases the rich distribution of multimodal clinical data supported by intuitive visualizations.

\subsection{Internal Datasets}

\subsection{Public Datasets}
To enhance MediScope’s generalization ability across a wide range of clinical tasks, we incorporate multiple publicly available medical datasets for pre-training. These datasets span both single-modality (text-only) and multimodal (image-text) scenarios and originate from various medical departments. Tab.~\ref{tab:Public_dataset} summarizes the source, modality, and size of the datasets used.

\begin{itemize}
  \item \textbf{Huatuo26M~\cite{huatuo26m}} is a large-scale Chinese medical textbook corpus containing approximately 26 million text entries covering fundamental clinical knowledge. It serves as a solid foundation for pretraining domain-specific language understanding.
  
  \item \textbf{CMtMedQA~\cite{CMtMedQA}} is a medical question-answering dataset. It is commonly used to evaluate models’ factual medical knowledge within Chinese clinical contexts.

 \item \textbf{MedQA~\cite{medQA}} is a medical question-answering dataset primarily in English, often used to assess models’ factual medical knowledge across various medical specialties.
  
  \item \textbf{MedMCQA~\cite{medmcqa}} is a large-scale multiple-choice question dataset spanning various medical specialties, commonly used for fine-tuned evaluation of medical reasoning.
  
  \item \textbf{PathVQA~\cite{pathvqa}} consists of pathology images paired with related questions, requiring both visual understanding and domain-specific medical knowledge to answer.
  
  \item \textbf{PMC-VQA~\cite{pmc-VQA}} is a medical visual question answering dataset derived from PubMed Central (PMC) articles. It contains over 220K examples across a wide range of clinical and biomedical topics.
  
  \item \textbf{SLAKE~\cite{slake}} is designed for complex medical VQA tasks, integrating radiological knowledge with visual and textual reasoning.
  
  \item \textbf{VQA-RAD~\cite{VQA-RAD}} is one of the earliest visual question answering datasets in radiology, built on real-world radiology images paired with physician questions, suitable for basic image-based QA research.
   \item \textbf{DermaVQA~\cite{dermavqa}} is a dermatology-focused VQA dataset with image-question pairs on skin conditions, supporting model training for dermatological diagnosis.
  
\end{itemize}

Together, these datasets form a comprehensive foundation for pretraining in both medical language understanding and vision-language tasks, enabling robust performance across various real-world clinical scenarios.

\begin{table}[t!]
\small
\centering
\caption{Medical pre-training data statistics and sources, all data are from real medical scenarios.}
\renewcommand{\arraystretch}{1.4}
\begin{tabular}{l l l l}
  \toprule
  \textbf{Dataset} & \textbf{Type} & \textbf{Modality} & \textbf{Size} \\
  \midrule
  Huatuo26M    & Textbook        & Text-Only    & 26M  \\
  CMtMedQA     & Q\&A         & Text-Only    & 68K \\
  MedQA     & Q\&A            & Text-Only    & 60K \\
  MedMCQA & Q\&A        & Text-Only    & 183K \\
  PathVQA    & Q\&A         & Image-Text    & 20K  \\
 PMC-VQA    & Q\&A         & Image-Text    & 227K \\
  SLAKE     & Q\&A            & Image-Text    & 14K \\
  VQA-RAD & Q\&A       & Image-Text    & 3K \\
  DermaVQA & Q\&A  & Image-Text    & 4K  \\
  \bottomrule
\end{tabular}
\label{tab:Public_dataset}
\end{table}

\subsection{Representative Medical Data Types in MediScope} 
The main text outlines the detailed distribution of various data types in the MediScope dataset, encompassing a rich array of diagnostic sources such as biochemical tests (15.9\%), ultrasound imaging (13.0\%), electronic medical records (10.6\%), blood and CRP tests (6.4\%), and multiple forms of medical imaging (6.0\%). To provide a more intuitive understanding of these heterogeneous data types, Fig.~\ref{fig:modality_distribution} presents visual examples from 12 representative medical sources, highlighting the diversity and comprehensiveness of the dataset.

\begin{figure}[t]
\begin{center}
\includegraphics[width=1\linewidth]{AnonymousSubmission/LaTeX/sup-fig/sup-show.pdf}
\end{center}
\vspace{-2mm}
   \caption{ Representative samples from the MediScope dataset, illustrating its extensive coverage across key medical data sources. \textit{Best viewed when zoomed in.}} 
\label{fig:modality_distribution}
\vspace{-2mm}
\end{figure}

As shown in Fig.~\ref{fig:modality_distribution}, MediScope integrates clinically significant data sources, including pathology reports, dermatology images, genetic testing records, prescriptions, cardiology reports, endoscopy images, etc. These modalities are rarely seen in existing public medical datasets. Collectively, they form a comprehensive multimodal dataset. This diversity enables broad coverage of real-world medical scenarios and establishes a solid foundation for developing robust medical multimodal models.


\section{Relative \emph{vs.} Absolute Evaluation Analysis}
To gain a clearer understanding of relative versus absolute evaluation strategies in the PulseMind Benchmark, we conduct a series of experiments comparing their scoring behavior and evaluation reliability. First, we analyze absolute scoring in distinguishing between different models. Second, we compare the reliability of both strategies by examining their consistency with evaluations made by human experts. The evaluated models include Lingshu~\cite{lingshu}, Gemini2.5-Pro~\cite{gemini-1.5}, O1~\cite{o1}, GPT-4o~\cite{gpt-4o}, InternVL3~\cite{internvl3}, and Qwen2.5-VL~\cite{qwenvl}.

\subsubsection{Analysis of Absolute Scoring Behavior.} To evaluate the distribution and discriminative capacity of the absolute scoring strategy, we rated seven models using a 5-point scale across four evaluation dimensions: Proactiveness (Proact.), Accuracy (Acc.), Usefulness (Use.), and Language Quality (Lang.).

\begin{table}[ht]
\small
\setlength{\tabcolsep}{2.8pt}
\centering
\begin{tabular}{lccccc}
\toprule
Model & Proact. & Acc. & Use. & Lang. & Avg. \\
\midrule
Lingshu & 3.60 & 4.15 & 4.05 & 4.58 & 4.10 \\
Gemini2.5-pro & 3.91 & 4.35 & 4.30 & 4.69 & \underline{4.31} \\
o1    & 3.70 & 4.43 & 4.14 & 4.60 & 4.22 \\
GPT-4o    & 3.48 & 4.06 & 3.97 & 4.52 & 4.01 \\
Qwen2.5VL-72B & 3.81 & 4.30 & 4.22 & 4.59 & 4.23 \\
InternVL3-78B & 3.74 & 4.36 & 4.16 & 4.71 & 4.24 \\
PulseMind        & 3.92 & 4.47 & 4.26 & 4.763 & \textbf{4.35} \\
\bottomrule
\end{tabular}
\caption{Absolute scores assigned to seven models across four evaluation dimensions. The top two results are highlight with \textbf{bold} and \underline{underlined} fonts, respectively.}
\label{tab:abs}
\vspace{-1em}
\end{table}
As shown in Tab.~\ref{tab:abs}, all models received relatively high and closely clustered scores, with average ratings ranging from 4.01 to 4.35. Such small score differences make it difficult to effectively differentiating the model performance.

\subsubsection{Consistency Analysis of Model and Human Expert Evaluations.}
To assess the reliability of the relative and absolute evaluation strategies, we examine their consistency with human expert judgments. For the absolute scoring strategy, each model is independently scored by both GPT-4o and human experts, and the results are compared to assess alignment. For the relative scoring strategy, we designate PulseMind as the baseline model and ask both GPT-4o and human experts to compare PulseMind against each of the other models. We then compute the win rate of PulseMind to measure consistency.

\begin{table}[ht]
\centering

\begin{tabular}{lcccc}
\toprule
\textbf{Model} & \textbf{M Abs} & \textbf{H Abs} & \textbf{M WR} & \textbf{H WR} \\
\midrule
Lingshu         & 4.10 & 3.0 & 98\% & 95\% \\
Gemini2.5-pro         & 4.31 & 4.2 & 54\% & 50\% \\
o1         & 4.22 & 2.8 & 89\% & 85\% \\
GPT-4o        & 4.01 & 4.0 & 94\% & 90\% \\
Qwen2.5VL  & 4.23 & 3.8 & 86\% & 88\% \\
InternVL3& 4.24 & 3.5 & 83\% & 80\% \\
PulseMind          & 4.35 & 4.5 & - & - \\
\bottomrule
\end{tabular}
\caption{Model performance evaluated by absolute scores and relative scores (win rate, \%). ``M” indicates the model-based evaluation, while ``H” refers to the human expert evaluation. ``Abs” represents absolute scores, and ``WR” denotes the win rate of the PulseMind model compared to other models.}
\label{tab:doctor}
\end{table}

The results show that relative scoring yields larger differences between models, making it more effective in distinguishing model capabilities. Moreover, it demonstrates strong consistency with human expert evaluations. In contrast, absolute scoring not only exhibits limited discriminative power but also shows a high degree of inconsistency with human assessments.

\section{Additional Experimental Results}
In this section, we present additional experimental results. First, we report dimension-wise evaluation results on the PulseMind Benchmark. Second, we provide performance results on subsets of the MMMU dataset.

\subsection{Dimension-wise Evaluation Results}

In the main text, we reported the overall win, tie, and loss distributions on the PulseMind Benchmark. Here, we further provide detailed results across four evaluation dimensions (proactiveness, accuracy, usefulness, and language quality) and two subsets (MedDiagnose and cMtMedQA-test). 

\begin{figure}[t]
\begin{center}
\includegraphics[width=1\linewidth]{AnonymousSubmission/LaTeX/sup-fig/sup_tianchi_4all.pdf}
\end{center}
\vspace{-2mm}
   \caption{Win rates across the four evaluation dimensions on the MedDiagnose dataset} 
\label{fig:MedDiagnose}
\vspace{-2mm}
\end{figure}

\begin{figure}[t]
\begin{center}
\includegraphics[width=1\linewidth]{AnonymousSubmission/LaTeX/sup-fig/sup_CMtMed_4all.pdf}
\end{center}
\vspace{-2mm}
   \caption{Win rates across the four evaluation dimensions on the CMtMedQA-test dataset.} 
\label{fig:CMtMedQA}
\vspace{-2mm}
\end{figure}

The corresponding results are presented in Figures~\ref{fig:MedDiagnose} and \ref{fig:CMtMedQA}. The results show that on both MedDiagnose and cMtMedQA-test subsets, the PulseMind model consistently achieves higher win rates than all compared models across all evaluation dimensions, demonstrating its strong performance in each aspect.




\subsection{Evaluation on the MMMU Subset}
Tab.~\ref{tab:mmmu_health} presents the performance of numerous models on the Health \& Medicine subset of the MMMU~\cite{mmmu} benchmark, across five subfields: Basic Medical Science (BMS), Clinical Medicine (CM), Diagnosis and Laboratory Medicine (DLM), Pharmacy (P), and Public Health (PH).

\begin{table}[htb]
\small
\centering
\renewcommand{\arraystretch}{1.1}
\resizebox{0.48\textwidth}{!}{%
\begin{tabular}{lccccc c}
\toprule
\textbf{method} & \textbf{BMS} & \textbf{CM} & \textbf{DLM} & \textbf{P} & \textbf{PH} & \textbf{average} \\
\midrule
internVL3              & 78.6 & 63.3 & 40.0 & 70.4 & 93.3 &  \underline{69.1}  \\
Qwen2.5vl-72B          & 71.4 & 66.7 & 46.7 & 74.1 & 73.3 & 66.4 \\
GPT-4o                 & 64.3 & 56.7 & 50.0 & 55.6 & 60.0 & 57.3 \\
o1                     & 71.4 & 63.3 & 50.0 & 44.4 & 60.0 & 57.8 \\
Gemini2.5-pro          & 63.3 & 66.7 & 43.3 & 30.0 & 43.3 & 49.3 \\
LLAVA-med              & 56.4 & 56.0 & 43.9 & 46.7 & 41.7 & 48.9 \\
HuatuoGPT-vision       & 64.6 & 62.5 & 45.9 & 54.1 & 44.2 & 54.3 \\
Lingshu & - & - & - & - & - & 62.3 \\
PulseMind   & 75.0 & 56.7 & 50.0 & 85.2 & 80.0 &\textbf{69.4}  \\
\bottomrule
\end{tabular}
}
\caption{Results on MMMU Health & Medicine. The top two results are highlighted using \textbf{bold} and \underline{underlined} fonts, respectively. References: LLAVA-med~\cite{llava-med}, HuatuoGPT-vision~\cite{huatuo-vision}.}
\label{tab:mmmu_health}
\end{table}

PulseMind achieves the highest overall average score of 69.4\%, slightly surpassing InternVL3 at 69.1\% and outperforming all other models by a notable margin. Beyond overall performance, PulseMind demonstrates balanced strengths across individual subfields. It ranks first in Pharmacy with a score of 85.2\%, showing strong understanding of medication-related content, and places second in Public Health with a score of 80.0\%, just behind InternVL3’s 93.3\%. In Diagnosis and Laboratory Medicine, it ties for the top score of 50.0\%. Its score in Clinical Medicine is 56.7\%, which is competitive with most baselines. In Basic Medical Science, PulseMind achieves 75.0\%, second only to InternVL3’s 78.6\%. These results highlight PulseMind’s consistent and well-rounded performance across various domains of medical knowledge.


\section{Evaluation Prompts}
The prompts used for evaluation on different datasets are illustrated in Fig.~\ref{fig:prompt}.

\begin{figure}[h]
\begin{center}
\includegraphics[width=1\linewidth]{AnonymousSubmission/LaTeX/sup-fig/sup-prompt.pdf}
\end{center}
\vspace{-2mm}
   \caption{The prompt formats of different question types.} 
\label{fig:prompt}
\vspace{-2mm}
\end{figure}




\section{Limitation}
One limitation of PulseMind lies in the high cost of dataset construction, as it requires extensive involvement of human medical professionals for validation and quality control. With the advancement of medical AI, we hope to gradually automate this pipeline to reduce the overall cost. In addition, PulseMind has yet to be extended to broader medical domains, such as long-term patient monitoring and drug discovery. In future work, we will focus on expanding the capabilities and applicability of medical models across a wider range of medical domains.